# OCTID: Optical Coherence Tomography Image Database

Peyman Gholami[1] , Priyanka Roy, Mohana Kuppuswamy Parthasarathy, Vasudevan Lakshminarayanan

*Theoretical & Experimental Epistemology Lab (TEEL), School of Optometry and Vision Science, University of Waterloo, 200 University Ave. West, Waterloo, ON, Canada N2L 3G1*

**Abstract**

Optical coherence tomography (OCT) is a non-invasive imaging modality which is widely used in clinical ophthalmology. OCT images are capable of visualizing deep retinal layers which is crucial for early diagnosis of retinal diseases. In this paper, we describe a comprehensive open-access database containing more than 500 high-resolution images categorized into different pathological conditions. The image classes include Normal (NO), Macular Hole (MH), Age-related Macular Degeneration (AMD), Central Serous Retinopathy (CSR), and Diabetic Retinopathy (DR). The images were obtained from a raster scan protocol with a 2mm scan length and 512x1024 pixel resolution. We have also included 25 normal OCT images with their corresponding ground truth delineations which can be used for an accurate evaluation of OCT image segmentation. In addition, we have provided a user friendly GUI which can be used by clinicians for manual (and semi-automated) segmentation.

## 1. Introduction

*Optical Coherence Tomography (OCT)*

Optical Coherence Tomography (OCT) is a non-invasive imaging modality which is of great importance in clinical ophthalmology [1]. OCT is utilized for the cross-sectional visualization of the retinal structures which is crucial for the early diagnosis of several pathologies that affect the retina and the optic nerve head, such as macular degeneration and macular edema [2]. Studies have shown that OCT can also be used to monitor the functionality of tissues [3] and blood flow within the retina as well as assess the level of blood oxygenation [4,5].

There are several SD-OCT databases available in the literature [6-11]. Most of these databases have several limitations, e.g., some have a limited number of images [9] and some only contain images from normal subjects [7]. However, the shape and the thickness of the retinal layers change based on different ocular conditions and OCT images are capable of visualizing these changes. Therefore, these databases are not able to provide such information. Some other databases have only focused on a specific ocular disease such as age-related macular degeneration [8] or glaucoma [9]. However, most of these databases differ with each other in terms of resolution, image quality, size, retinal field area covered and other properties of the images and due to these differences, comparing them is a challenging task. The available OCT retinal databases are discussed in detail in a recent review paper [10]. A database which contains different ocular conditions with the same image characteristics would be ideal for researchers in this field.

---

[1] email:pgholami@uwaterloo.ca

Such databases are of practical importance for evaluating the performance of different segmentation and classification algorithms. Moreover, such an organized database helps researchers to create more efficient methods for computer-aided identification of ocular diseases. We have created an open-access OCT image database which includes high-resolution OCT images with different retinal related diseases. We also include manual segmentation by an experienced clinician of a set of images as the ground truth delineations and these can be used as the gold standard for the evaluation of different image analysis methods.

## 2. Structure of the Database

*Database Images properties*

The database contains OCT images categorized into different diseases and includes high-resolution jpeg images that can be downloaded as zip files. The database consists of more than 500 spectral-domain OCT volumetric scans, consisting of four categories: Normal (NO), Macular Hole (MH), Age-related Macular Degeneration (AMD), Central Serous Retinopathy (CSR), and Diabetic Retinopathy (DR).

The images were from a raster scan protocol with a 2mm scan length, containing 512x1024 pixels, captured using a Cirrus HD-OCT machine (Carl Zeiss Meditec, Inc., Dublin, CA) at Sankara Nethralaya (SN) Eye Hospital, Chennai, India. In each volumetric scan, a fovea-centered image was selected by an experienced clinical optometrist (MKP). The axial and transverse resolution was 5 μm, 15 μm respectively (in tissue). The optical source was a Superluminescent diode (SLD) with 840 nm wavelength. For more information concerning the specification of the images can be found in the device user manual [12]. The images were then resized to a 500x750 pixel size. The pathologies were diagnosed by clinicians and image class labeling was done based on the diagnosis of retinal clinical experts at SN hospital.

Each of the datasets contains images of different stages for each of the diseases, including less severe, medium severe, and more severe stages. This wide range of stages in each set would be ideal for the researchers who want to test the sensitivity of their technique for each of the diseases or want to evaluate the repeatability of their system among different stages. The severity of the images is randomized over the dataset and there is no order on the distribution of images in each disease.

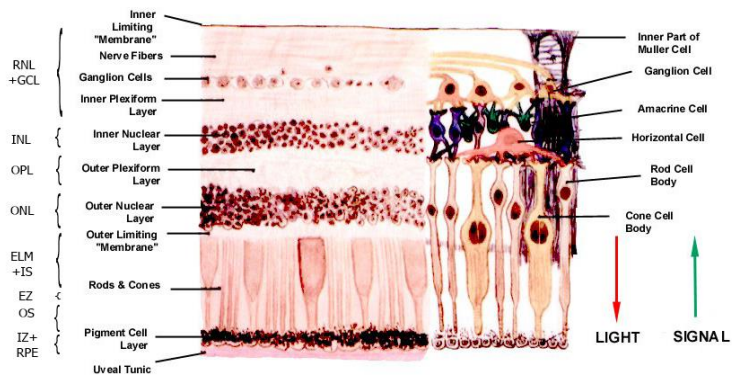

Figure 1. The structure of the normal retina with different retinal

*Pathological conditions*

In this section, we provide a brief discussion about each of the disease groups. These diseases account for over 25% of severe visual impairment causes [13,14] and affect different retinal layers. Figure 1 shows the structure of the normal retina with different layers indicated on it. Figure 2 presents the OCT image of a healthy normal retina. These diseases are described in detail in the literature (e.g., [15,16]). In what follows, we present some general information about each disease and their symptoms. We also provide two sample OCT images for each type of the diseases, one showing a less severe stage, and one image presenting a more severe stage of the disease:

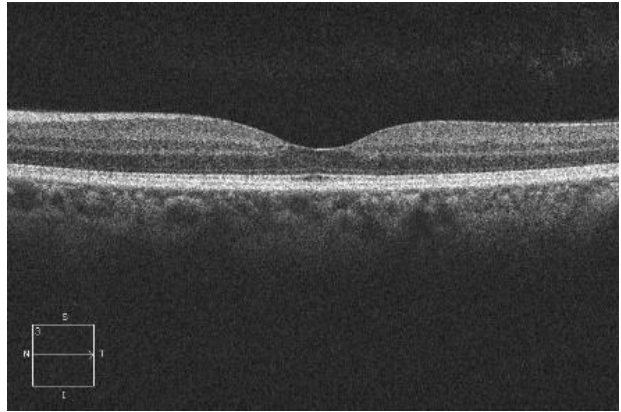

Figure 2. Healthy normal retinal OCT image

• Diabetic retinopathy (DR), is one of the most leading causes of blindness in adults over the age of 65 years in the United States [17]. Due to diabetes, retinal blood vessels can break down, leak or become blocked. Diabetic macular edema can occur in any stage of DR and cause several changes to the morphology of retinal layers, including increased retinal thickness and focal retinal detachments. This is due to intra-retinal fluid accumulation and is indicated by reduced intra-retinal reflectivity. Other visible changes on OCT images include abnormal foveal contour, changes in the central macular thickness, and intra-retinal hyper-reflective exudates. [17].

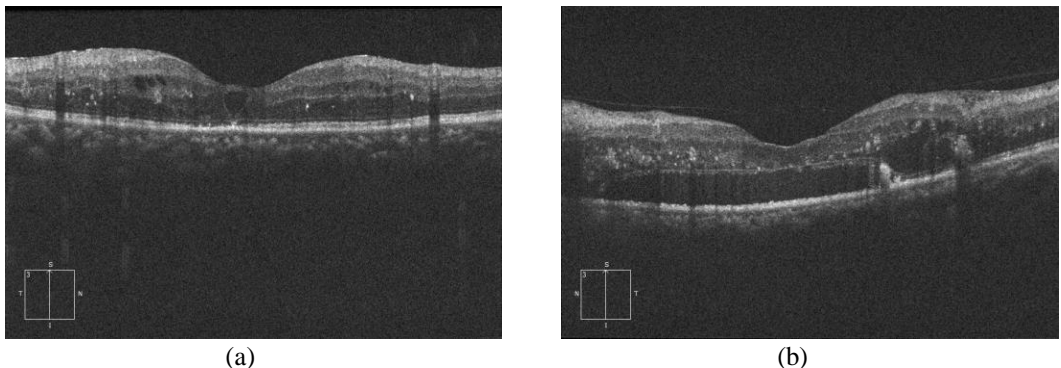

(a) (b)
Figure 3. Sample DR images. (a) less severe stages (b) severe stages of the disease

• Age-related Macular Degeneration (AMD) is the leading cause of severe visual impairments and vision loss in Western countries. AMD affects the part of the retina responsible for sharp central vision, namely the macula. The OCT image for AMD usually appears with thickening of Retinal Pigment Epithelium (RPE) layer which results in the elevation of the photoreceptor layer and thickened, irregular, and disorganized retinal layers. Also, it usually causes fluid accumulation beneath the fovea and in some cases, the detachment of the RPE layer can occur [17].

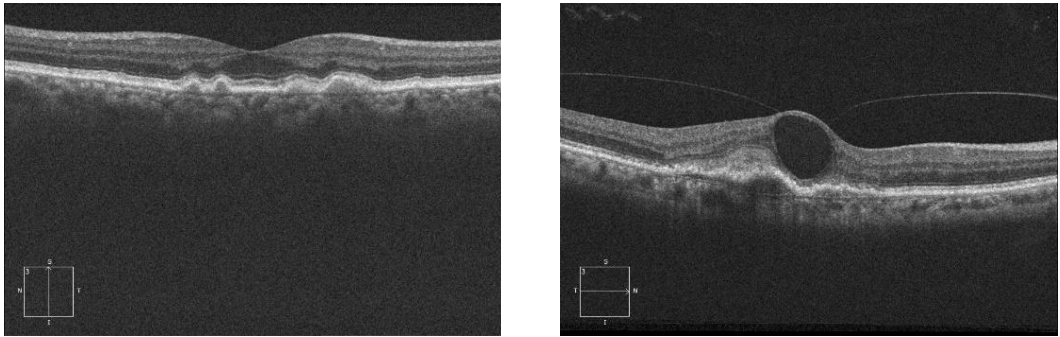

(a)                                              (b)
Figure 4. Sample AMD images. (a) less severe stages (b) severe stages of the disease

• Macular Hole (MH) is another common age-related eye disease and causes a decrease in visual acuity. MH retinal defects can vary from small hypoflective breaks to full-thickness intra-retinal spaces in the OCT images. Other signs include thickened edges with cavities of reduced reflectivity, macular edema, and fluid accumulation [17]. Furthermore, using OCT images one can understand the etiology of the MH disease.

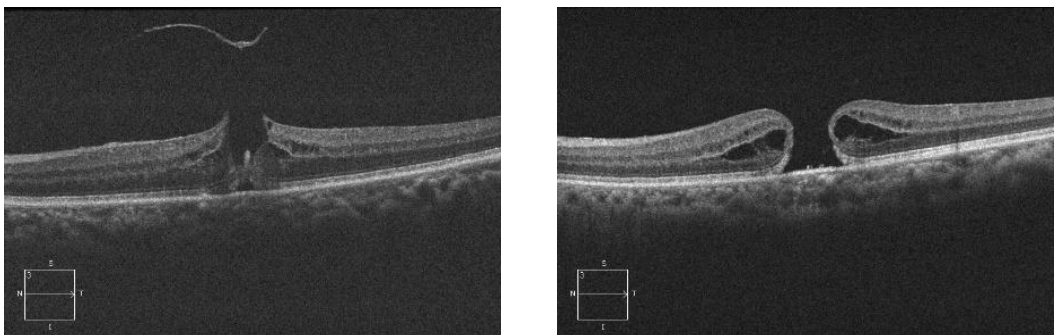

(a)                                              (b)
Figure 5. Sample MH images. (a) less severe stages (b) severe stages of the disease

• Central Serous Retinopathy (CSR) is a common disease of the middle age (third and fourth decade) that causes visual distortions and a decrease in visual acuity. It is characterized by focal detachments of the retina and the RPE layer due to the leakage of fluid into the retina through an RPE defect. On OCT, CSR appears as hypo reflective spaces at the sub-retinal and sub-RPE levels with occasional hyper-reflective deposits sub-retinally. OCT also helps in differentiating CSR from other confounding choroidal diseases [18].

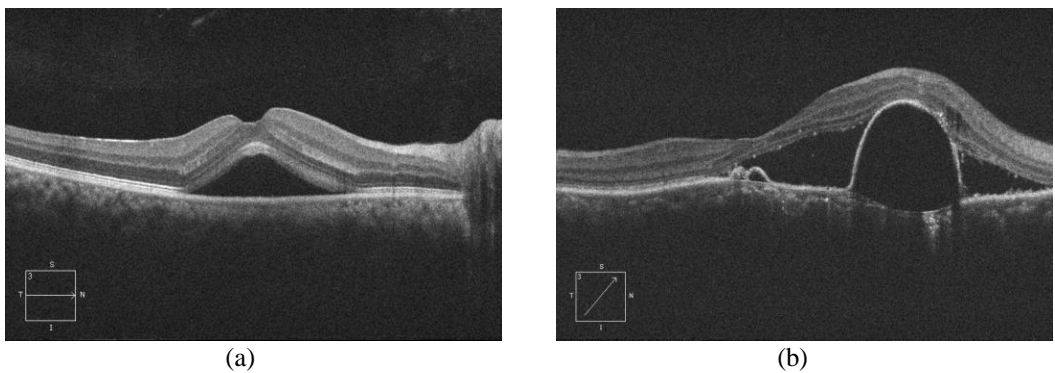

Figure 6. Sample CSR images. (a) less severe stages (b) severe stages of the disease

*Manual segmentation*

We also included 25 normal images along with the corresponding ground truth delineations, which have been determined by an expert. The ground truth locations are added as MAT files in the database. The manual segmentation was done by an experienced clinician (MKP). In addition, to aid the novice user or clinician, we have developed an easy to use GUI (Fig. 7) for use with the dataset. The clinician will have the ability to select the name of the boundary from the listing box. He/she can choose to re-segment the whole of that boundary manually by clicking on the "manual" button or re-segment only the erroneous portions within the boundaries by clicking on the "semi-auto" button. After re-segmentation of all the layers is completed by the clinician, the "Exit" button is clicked, when it prompts the user whether the user wanted to save the new segmentation or not.

*Database structure*

Our database consists of 102 MH, 55 AMD, 107 DR, and 206 NO retinal images. The database is available to the public at:
http://doi.org/10.3886/E108503V1 and at
https://dataverse.scholarsportal.info/dataverse/OCTID

Each of the different classes is categorized as a separate dataset and is available with a unique DOI link. The images can be viewed and downloaded separately or entirely as a zip file.

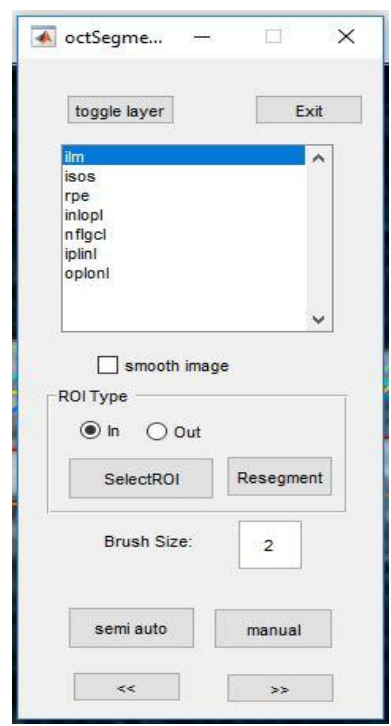

Figure 7. The Graphical user interface (GUI) for manual segmentation

## 3. Future Work and Conclusion

In this paper, we introduced an open access OCT image database (OCTID). This possibly the largest OCT retinal image database that is currently available for researchers that provides a comprehensive collection of high-resolution images and is categorized based on pathological conditions. The ultimate goal for creating this database was to facilitate public access to OCT images in order to help researchers worldwide who want to develop computer-aided algorithms for the analysis of OCT images and might not have access to a sufficient number of good quality images. This is also important for researchers in deep learning developing neural networks for automated classification of diseases [19, 20]. Moreover, one of the other major potential uses of this dataset is for image segmentation studies. The majority of the current OCT image segmentation studies have mainly focused on delineating the normal retinal layers [21]. However, as other studies have shown, the segmentation of retinal layers with pathological conditions is of much more importance in preventing vision loss and this dataset can provide the means for such studies. Therefore, by removing such a barrier, we hope that more efficient techniques/algorithms can be created, tested and validated which can be of use for accurate diagnosis.

In addition, we will further work on improving this database in order to create a much more comprehensive data set. We also have access to a number of images (greater than 100,000). We are in the process of examining and categorizing these images and incorporating them. Therefore, by establishing such a huge database, we can build novel neural networks and train them using these images. It is expected that by using such a large database, the performance of the neural network will increase dramatically, and we can create a highly reliable identification and classification technique for automated diagnosis of ocular diseases.

**Acknowledgment**

The authors would like to acknowledge Sankara Nethralaya (SN) eye hospital, Chennai, India, and in particular Dr. Muna Bhende and Ms. Girija for providing the images and diagnosing pathological conditions. This research is supported by an NSERC discovery grant to VL. MKP was supported by a Faculty for the Future fellowship from Schlumberger Foundation,